\documentclass[letterpaper, 10 pt, conference]{ieeeconf}  %

\IEEEoverridecommandlockouts                              %

\usepackage{graphicx} %
\usepackage{amsmath} %
\usepackage{amsfonts}
\usepackage{gensymb}
\usepackage{hyperref}
\usepackage[table]{xcolor}    %

\newcommand{\sysName}{FORGE}

\title{\LARGE \bf
\sysName: Force-Guided Exploration for Robust Contact-Rich Manipulation under Uncertainty
}
\author{
Michael Noseworthy$^{1}$, Bingjie Tang$^{2}$, Bowen Wen$^{3}$, Ankur Handa$^{3}$,  Chad Kessens$^{4}$, Nicholas Roy$^{1}$ \\ Dieter Fox$^{3}$, Fabio Ramos$^{3}$, Yashraj Narang$^{3}$, Iretiayo Akinola$^{3}$ 
\thanks{$^{1}$MIT $^{2}$USC $^{3}$NVIDIA $^{4}$DEVCOM ARL.} 
}

\begin{document}

\maketitle
\thispagestyle{empty}
\pagestyle{empty}

\begin{abstract}

We present \sysName, a method for sim-to-real transfer of force-aware manipulation policies in the presence of significant pose uncertainty. During simulation-based policy learning,
{\sysName} combines a \emph{force threshold} mechanism with a \emph{dynamics randomization} scheme to enable robust transfer of the learned policies to the real robot.
At deployment, {\sysName} policies, conditioned on a maximum allowable force, adaptively perform contact-rich tasks while avoiding aggressive and unsafe behaviour, regardless of the controller gains.
Additionally, {\sysName} policies predict task success, enabling efficient termination and autonomous tuning of the force threshold.
We show that {\sysName} can be used to learn a variety of robust contact-rich policies, including the forceful insertion of snap-fit connectors.
We further demonstrate the multistage assembly of a planetary gear system, which requires success across three assembly tasks: nut threading, insertion, and gear meshing. Project website: \href{https://noseworm.github.io/forge/}{https://noseworm.github.io/forge/}
\end{abstract}

\section{Introduction}
We are interested in developing \emph{sim-to-real} techniques for learning assembly primitives (e.g., low-clearance insertion or nut-threading).
Over the past decade, sim-to-real techniques have led to advances in dexterous manipulation and legged locomotion \cite{akkaya2019solving, handa2022dextreme, tan2018sim, hwangbo2019learning}.
However, similar results have only recently been achieved for robotic assembly, which requires efficient and accurate simulation of the detailed, low-clearance parts \cite{narang2022factory, yoon2022fast, tang2023industreal, schoettler2020meta, kozlovsky2022reinforcement, beltran2020variable}.
Even with these advances, successful sim-to-real deployment remains challenging for contact-rich tasks.

\begin{figure}
    \centering
    \includegraphics[width=0.7\linewidth]{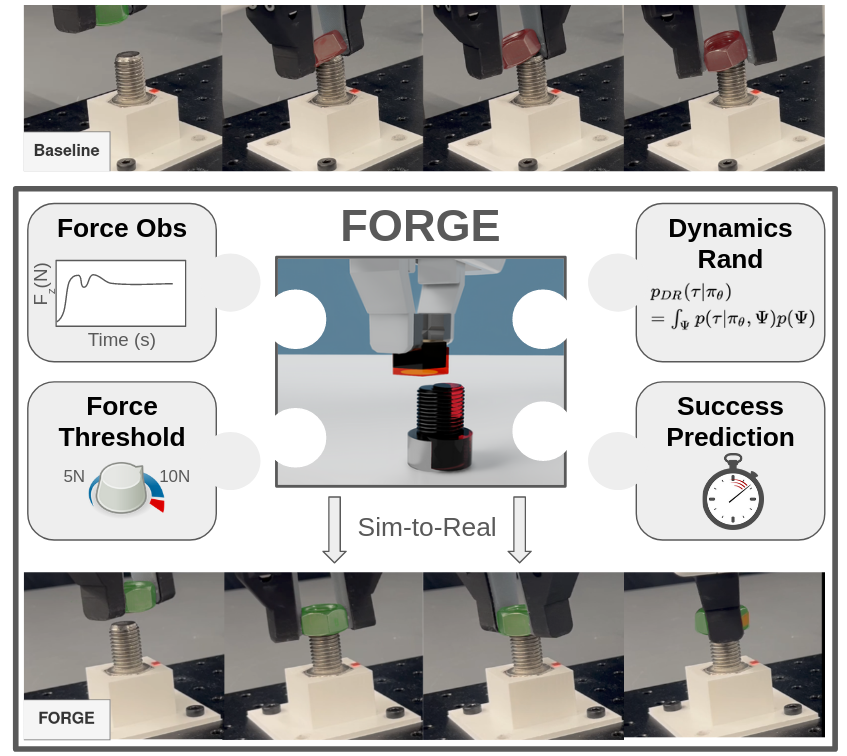}
    \caption{\label{fig:main} {\sysName} uses force feedback to learn search behaviours for contact-rich tasks with position estimation uncertainty. It combines \textit{dynamics randomization}, a \textit{force threshold}, and \textit{success prediction} for robust sim-to-real transfer. The resulting policies are \emph{safe} and \emph{efficient} (bottom) compared to aggressive baseline policies that cause parts to slip (top).}
    \vspace{-1.5em}
\end{figure}

Naively, policies can be too aggressive, leading to catastrophic part slip or damage that makes the task difficult or impossible to complete (see Figure \ref{fig:main}).
This is particularly pronounced when there is pose uncertainty and search behaviours that rely on contact are necessary \cite{chhatpar2001search, jin2021acc}.
The required contact between parts can lead to undesirable outcomes if the forces are too high.
Heuristic approaches, such as spiral search \cite{chhatpar2001search, van2018comparative}, can limit the applied force but these approaches are task-specific and can be inefficient.

Reinforcement learning offers a general paradigm for developing more flexible search behaviours.
However, previous works typically rely on additional procedures to ensure policies are deployed safely with desirable force profiles.
For example, policies trained using the \emph{IndustReal} framework \cite{tang2023industreal} do not observe or adapt to contact forces.
Instead, as our experiments show, forceful behaviour is determined by careful controller design and gain tuning.
Other works provide methods to optimize or adapt gains online \cite{zhang2023efficient,zhang2024bridging}.
Importantly, the desired force profile of a policy depends on the task at hand.
For example, threading a nut may fail if the applied force is too high, while a snap-fit connector might require large forces to ensure proper insertion.
Therefore, it is important to have simple and efficient methods to tune the policy's force profile.

In this work, we propose {\sysName}: a framework for developing force-aware sim-to-real policies for assembly tasks.
{\sysName} policies, trained solely in simulation, use external force observations to achieve efficient and gentle behaviour.
Additionally, policies are trained without precise knowledge of part poses, leading to emergent search behaviours that are robust to significant levels of pose uncertainty. 

{\sysName} has two complementary components to ensure policies are robust to contact.
First, we propose to condition policies on a \emph{force threshold} that should not be exceeded during task execution.
Second, policies are trained to maintain this threshold under a wide range of dynamics randomizations (we randomize \emph{robot}, \emph{controller}, and  \emph{part} properties).
Together, these components result in policies that can modulate their actions to achieve a force profile that respects the interpretable scalar force threshold. 
By randomizing this threshold during training, we are able to tune it at deployment time without retraining the policy.

For tasks with high enough clearance, a small force threshold is sufficient and no tuning is necessary.
However, tasks that require significant force to succeed (e.g., snap-fit connectors) may fail if the threshold is set too low.
When the required force is not known a priori, we present an automatic tuning procedure which leverages a notion of \emph{success prediction}.
Based on the outcome of a policy execution, the threshold can be iteratively adjusted for future trials.
To automate this tuning, {\sysName} policies are trained to predict whether an episode succeeded or failed.
We validate this procedure by showing successful sim-to-real transfer on a snap-fit connector requiring $15N$ for insertion.

Furthermore, we show how success prediction can lead to more efficient policy termination.
Standard practice in \emph{sim-to-real} assembly is to execute policies for a fixed duration \cite{tang2023industreal} which can lead to premature termination or delays. 
Instead, the policy can terminate when it believes it is in a successful state.
We show that success prediction, also trained in simulation, robustly transfers to the real world and does so more reliably when using force observations \cite{huang2022training}.

In summary, our contributions are:
\begin{enumerate}
\item \textbf{A method to specify maximum allowable contact-force} during policy execution. This results in policies that exhibit safe search behaviour even with significant levels of position estimation error (up to $5mm$). 
\item \textbf{A dynamics randomization scheme} that reduces tuning to an interpretable scalar force-threshold parameter (instead of controller gains).
\item \textbf{A method for success prediction} that enables automatic force-threshold tuning and efficient policy termination, reducing delay times up to $66\%$.
\item \textbf{A demonstration of multi-part assembly} of a planetary gearbox requiring a diverse set of skills, including the challenging task of fastening nuts and bolts. %
\end{enumerate}

Results are shown for over $1000$ real-world trials and multiple tasks. We plan to release the code with the paper.

\section{RL for Contact-Rich Assembly}
We want to learn policies for tasks with tight tolerances and detailed geometry.
We first describe the problem formulation before introducing {\sysName} in the next section.

\subsection{Assembly Tasks}
\label{sec:background-tasks}

\begin{figure*}[t]
    \centering
    \vspace{1.0em}
    \includegraphics[width=0.9\linewidth,trim={0cm 0cm 0cm 1cm},clip]{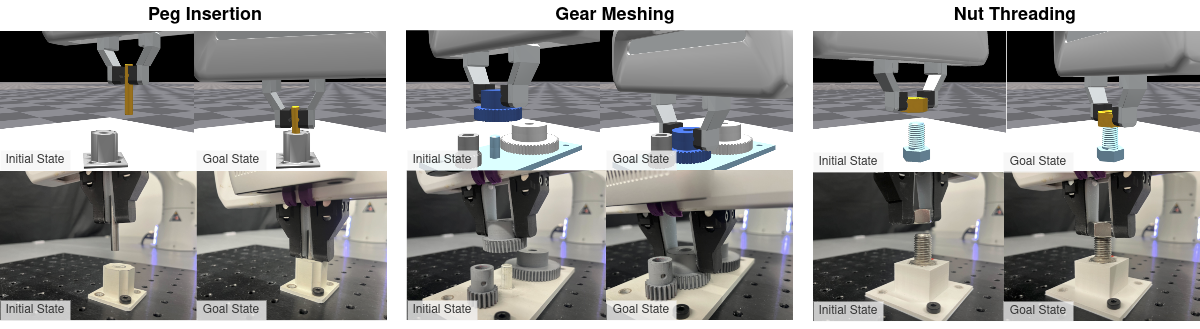} 
    \vspace{-1em}
    \caption{\label{fig:tasks} {\sysName} is evaluated on three tasks from \emph{Factory} \cite{narang2022factory}: Peg Insertion, Gear Meshing, and Nut Threading. Each task is trained solely in simulation (top) and transferred directly to the real robot (bottom).}
    \vspace{-1.0em}
\end{figure*}

Each task involves mating two parts: one grasped and another fixed to the workspace.
For our main evaluations, we consider all three tasks from \emph{Factory} \cite{narang2022factory} and demonstrate the first sim-to-real transfer for threading a small M16 nut (see Fig. \ref{fig:tasks}). 
We also consider forceful snap-fit insertion and multi-step assembly in Sec. \ref{sec:results-ft} and Sec. \ref{sec:results-gearbox} respectively.

\textbf{Peg Insertion:} A round peg with $8mm$ diameter needs to be inserted into a socket with $0.5mm$ diametrical clearance. 

\textbf{Gear Meshing:} A gear needs to be inserted onto a peg with $0.5mm$ clearance. Other gears are present and the teeth of adjacent gears must be aligned for successful meshing.%

\textbf{Nut Threading:} Instead of fully lowering a nut onto a bolt as in \emph{Factory}, we define the \emph{nut threading} task as successfully threading the nut such that it cannot be lifted by a vertical motion (we find lowering by a quarter-thread is sufficient).
Because our robot has joint limits, and to prevent the need to regrasp, we assume the nut and bolt are initially oriented\footnote{We leave the more challenging, yet realistic, scenario involving completely unobserved thread orientation to future work.} such that success can be achieved with a single revolution of the wrist joint.
We consider nuts with a relatively small size (M16) compared to previous sim-to-real work (M48) \cite{son_sim--real_2020}.
A successful search behaviour will resolve lateral uncertainty and place the nut on the bolt before rotating the wrist (otherwise the threads may not mesh).

\subsection{POMDP Formulation}
\label{sec:background-pomdp}

We formulate our problem as a \emph{Partially Observable Markov Decision Process} (\emph{POMDP}) \cite{kaelbling1998planning, jiang2024transic}.%
The goal is to learn a policy, $\pi_{\theta}(a_t|o_1,\ldots,o_t)$, that maximizes the expected return:
\begin{equation} \label{eq:background-objective}
J(\pi_{\theta}) = \mathbb{E}_{\tau \sim p(\tau |\pi_{\theta}, \Psi)} [  \Sigma_{t=0}^{\infty} \gamma^t r_t ]
\end{equation}
where $\tau = (s_0, a_0, o_0, s_1, a_1, o_1, \ldots)$ is the trajectory of states, actions, and observations resulting from the robot following policy $\pi_{\theta}$. 
Below, we further specify the components of the POMDP for contact-rich tasks.

\textbf{States ($\mathcal{S}$)}: 
A state, $s_t \in \mathcal{S}$ consists of the pose and velocities of the end-effector (EE), fixed part, and held part: $p^{ee}, p^{fixed}, p^{held} \in SE(3)$ and $v^{ee}, v^{held} \in \mathbb{R}^6$.%
We also include the contact force experienced by the end-effector, $F^{ee} \in \mathbb{R}^3$, and time-invariant information about the dynamics properties of the robot, controller, and parts (e.g., mass or joint-friction): $\Psi = (\psi_{robot}, \psi_{control}, \psi_{parts})$.

\textbf{Observations ($\Omega$)}:
As it is difficult to accurately estimate the full state, all our policies observe: 
\begin{itemize}
\item Noisy EE pose and velocity: $\hat{p}^{ee} \in SE(3)$, $\hat{v}^{ee} \in \mathbb{R}^6$
\item Estimated contact force: $\hat{F}^{ee} \in \mathbb{R}^3$
\item Noisy estimate of the fixed part's pose: $\hat{p}^{fixed} \in SE(3)$
\end{itemize}
We do not include pose or velocity of the held part because it can move in the gripper and be difficult to track.%
Likewise, we do not observe $\Psi$, but include the previous action, $a_{t-1}$, to help infer dynamics. 
See 
App. \ref{app:randomization} for noise models.

\textbf{Actions ($\mathcal{A}$)}: Control targets for a task-space impedance controller \cite{tang2023industreal,martin2019variable}.
As in previous work \cite{narang2022factory, tang2023industreal}, we assume all parts are in an upright orientation.
Thus it is sufficient for the policy to only have control authority over the ($x, y, z, yaw$)-dimensions: $a_t \in \mathcal{A} = \mathbb{R}^4$.

\textbf{Transition Function ($T_{\Psi}: \mathcal{S} \times \mathcal{A} \rightarrow \mathcal{S}$)}:
$T$ is parameterized by the dynamics parameters, $\Psi$ and is specified using the \emph{IsaacGym} \cite{makoviychuk2021isaac} simulator.
The sim-to-real gap comes from the mismatch between $\Psi^{sim}$ and $\Psi^{real}$.

\textbf{Observation Function ($O: \mathcal{S} \times \mathcal{A} \rightarrow \Omega$)}:
The position of the fixed part is assumed to have up to $5mm$ error.
Gaussian noise is assumed for each of the other observations.%

\textbf{Reward Function ($R: \mathcal{S} \times \mathcal{A} \rightarrow \mathbb{R}$)}:
Each task is described using a keypoint reward: $R_{kp}(p^{fixed}, p^{held}_t)$ \cite{narang2022factory, allshire2021transferring}, which is modified to account for small, threaded geometries (see App. \ref{app:reward} for more details).
We also add two discrete \emph{bonus} rewards that are given when important phases of the tasks are reached:
once the held part is centered on top of the fixed part and once the task is successful:
\vspace{-0.5em}
\begin{equation}
    R_{bonus}(p^{fixed}, p^{held}_t) = \mathbb{I}_{place} + \mathbb{I}_{success}.
    \vspace{-0.5em}
\end{equation}
We found the bonuses led to more robust learning when there is significant pose uncertainty.

\section{FORGE: Robust Search under Uncertainty}
{\sysName} uses on-policy RL to learn search behaviours in simulation.
A \emph{force threshold} (Sec. \ref{sec:forge-threshold}) and \emph{dynamics randomization} (Sec. \ref{sec:forge-dr}) are introduced for robust sim-to-real transfer.
{\sysName} also introduces \emph{success prediction} (Sec. \ref{forge-term}) for efficient termination and force threshold tuning. 

\subsection{Force Threshold}
\label{sec:forge-threshold}

During policy execution, excessive force can cause parts to slip or become damaged.%
Although it may be possible to recover from small amounts of slip with the right sensors (e.g., wrist camera or tactile), we prefer to avoid these scenarios.
Instead, we propose to condition the policy on a \emph{force threshold}, $F_{th}$: $\pi(a|o, F_{th})$.
During training, the policy is penalized if the contact force, $F^{ee}_t$, exceeds the threshold.
Concretely, we add an additional term to the reward function:
\vspace{-1em}
\begin{equation}
    R_{contact\_pen}(F^{ee}_t) = - \beta*\max(0, ||F^{ee}_t|| - F_{th}).
\end{equation}
Note this is related to \cite{beltran2020variable} which conditions the policy on a \emph{desired force} instead of an \emph{excessive force}.

At deployment time, $F_{th}$ can be set or tuned based on task requirements.
Most of the tasks we consider have positive clearance and do not require much force to succeed.
A relatively low force threshold is sufficient to prevent slip and tuning the threshold is not necessary.
For forceful insertion, nominal insertion forces are often speficied on part datasheets and the threshold should be higher than this value.
We also present an automatic tuning procedure in Sec. \ref{forge-term}.

\subsection{Dynamics Randomization}
\label{sec:forge-dr}

To successfully deploy policies trained in simulation, it is important that the trajectory distribution experienced during training is similar to what it would be when deployed: $p(\tau^{real}|\pi_{\theta}, \Psi^{real}) \approx p(\tau^{sim}|\pi_{\theta}, \Psi^{sim})$.
The difference between these distributions is usually referred to as the \emph{sim-to-real gap}.
This gap is usually handled by (1) system identification (Sys-ID) \cite{ljung1998system} or (2) dynamics randomization (DR) \cite{beltran2020variable, peng_sim--real_2018}.
The goal of Sys-ID is to tune $\Psi^{sim}$ to be close to $\Psi^{real}$.
This itself is a complicated tuning procedure that may need to be redone for every new set of parts.

Instead, we follow the DR approach which learns policies that are \emph{robust} to a wide range of dynamics parameters.
Concretely, we optimize a version of Eq. \ref{eq:background-objective} where:
\vspace{-0.5em}

\begin{equation}
\tau \sim p_{DR}(\tau|\pi_{\theta}) = \smallint p(\tau|\pi_{\theta}, \Psi) p(\Psi) d\Psi.
\end{equation}
The integral is approximated with Monte Carlo samples from a randomization distribution.
We now describe the variables that are randomized (see App. \ref{app:randomization} for values).

\textbf{Controller Randomization}: The controller has a large impact on contact forces. 
This work uses impedance-control where applied forces are computed as:
\vspace{-0.5em}
\begin{align}
    p^{targ}_t &= clip(combine(a_t, p^{fixed}), \lambda), \\
    F^{targ} &= k_p (p_t^{targ} - p_t^{ee}) - k_d v_t^{ee}.
\end{align}
First, the policy outputs a relative pose, $a_t$, which is applied to the fixed part's pose to get an absolute target pose, $p^{targ}_t$.
This pose is clipped by an action scale, $\lambda$, to ensure that the target is not too far from the EE's current pose.
As in previous work, we use critically damped gains to ensure stable controllers: $k_d = 2\sqrt{k_p}$ \cite{beltran2020variable, zhang2023efficient, oren2021iros}.
The controller thus depends on two parameters which govern how much force can be commanded: $\lambda \times k_p$.
We randomize both quantities so that the range of maximum commandable forces is in $[6.4, 20.0]N$.
Note that the control parameters are not included in the observations, so the policy must adjust its behavior based on force measurements.
This reduces the policy's dependence on a particular controller implementation.

Controller tuning \cite{tang2023industreal} or optimization \cite{zhang2023efficient} is a costly and often complex procedure.
Randomization \cite{peng_sim--real_2018} has the additional benefit that the policy is robust to a range of control parameters, greatly simplifying deployment.

\textbf{Part Randomization}: As parts slide against each other, material friction will affect lateral forces. To ensure policies can work across a range of materials, we randomize part mass and friction \cite{peng_sim--real_2018, apolinarska2021con, hebecker2021aim}.

\textbf{Robot Dynamics Randomization}: 
Due to phenomena such as joint friction \cite{peng_sim--real_2018, petrea2021frankaforce}, the applied force may be smaller than the commanded force.
We implement a simple way to account for this: inducing a randomized \emph{dead-zone} in simulation. 
Each episode, a dead-zone is selected for each dimension, $F^{DZ}_i$, where commanded forces below this value are clamped to zero:
$|F^{applied}_i| = \max(0, |F^{targ}_i| - F^{DZ}_i)$.
This enables the policy to increase its target which can help apply more force when needed or reduce steady-state error.

These randomizations lead to a policy that is robust to a wide range of dynamics parameters.
Combined with the force threshold, the policy can modulate its actions to achieve safe interaction.
For example, with higher gains, the policy will output smaller actions to limit the contact force.

\subsection{Success Prediction}
\label{forge-term}

Although success is clearly defined in simulation where we have access to noiseless poses, it is difficult to reliably predict in the real world \cite{huang2022training}.
Consider the nut-threading task, where the distance between a successfully threaded nut and a loose nut is a fraction of a millimeter.
We propose to train a success predictor which can robustly transfer from sim-to-real.
Concretely, we share the weights of the policy network with the success predictor by expanding the action space of the policy to include an early termination action: $a^{ET}_t \in [0, 1]$. 
To train the policy to output the correct action, we include an early termination penalty, $R^{ET}_t$, which penalizes incorrect success predictions:
\begin{equation}
    R^{ET}_t(a_t, y_t) = -|a^{ET}_t - y_t|, 
\end{equation}
where $y_t$ is the true success label at time $t$.
This reward can also encourage behaviours that elicit the underlying success state (e.g., pull upwards on the nut to check if it is threaded).

\textbf{Early Termination:} Efficient termination is a desirable property for industrial applications where cycle times matter.
We want the policy to terminate as soon as the task has succeeded and no sooner. 
During training, episodes are executed for the maximum length.
At deployment, a confidence threshold, $p_{term}$, can be used to terminate the episode: $a^{ET}_t > p_{term}$. See App. \ref{app:early-term} for analysis on the performance trade-offs for choosing $p_{term}$.

\textbf{Force Threshold Tuning:} 
For forceful insertion tasks, we may not know how much force is required.
Consider snap-fit connectors which require deformation.
The required force depends on material properties that may be unknown and could change with extended use. 
To tune the force threshold, we leverage success prediction.
Conservatively, we start with a low threshold of $7.5N$, which helps avoid slip and damage.
If policy execution reaches a timeout before success is predicted, we increase the threshold and try again.
This can be done automatically, without manual resets, until success occurs.

\section{Experiment Setup}
\subsection{Robot System}
\label{sec:exp-robot}

We use a \emph{Franka Panda} robot with the \emph{FrankaPy} \cite{zhang2020modular} library for impedance control.
All policies send control targets at $15Hz$ while the controller operates at $1000Hz$.
The Panda has joint-torque sensing, which is projected to EE-frame forces when needed by the policy  \cite{petrea2021frankaforce}.
Alternatively, a force-torque sensor could be used.

For the majority of our experiments, we calibrate the poses of each fixed object and artificially add noise.
This allows us to analyze performance under known levels of position estimation error.
The calibration is done by guiding the arm to a successful pose for the respective task from which a nominal initial pose can be backed out.
Unless otherwise reported, our real experiments use the same initial state randomization as in simulation (see App. \ref{app:randomization}).
For our last experiment, we assemble a planetary gear box (Sec. \ref{sec:results-gearbox}) using the perception system from \emph{IndustReal} \cite{tang2023industreal} (see App. \ref{app:gearbox} for more details).

\subsection{Policy Training}

\textbf{Simulator}: All policies are trained using the \emph{Factory} simulation methods within IsaacGym \cite{narang2022factory}.
In simulation, we have access to external contact forces experienced by the end-effector (akin to what we have access to on the Panda).
Noisy forces are used as policy input, whereas ground-truth forces are used to compute the excessive-force penalty.
We use recurrent PPO \cite{schulman2017proximal} with asymmetric actor-critic \cite{pinto2017asymmetric} to handle partial observability. %
Details on initial state and observation randomization can be found in App. \ref{app:randomization}.

\textbf{Checkpoint Selection}:
For all tasks and models, we train three policies with separate random seeds.
On the real-robot, results are averaged across the three policies.

\textbf{Observation and Action Frames}:
For generalization across the workspace, we assume actions and observations are relative to the fixed part.
Specifically, the policy outputs a $4D$ relative transform from the tip of the fixed part (we assume upright parts). 
The control target is computed from the fixed part's pose estimate and the relative pose from the policy.
The policy output is bounded, limiting the operational volume of the end-effector (targets can be up to $5cm$ away in all directions).
Similar to the action space, all position observations are relative to the tip of the fixed part.

\subsection{Baselines and Ablations}

We compare FORGE to two baselines:

\textbf{IndustReal \cite{tang2023industreal}:} Policies trained using the IndustReal framework do not have velocity or force observations. A full description of the differences can be found in App. \ref{app:industreal}.
The PLAI parameters from IndustReal are set to achieve similar maximum forces to what FORGE policies can command.

\textbf{Baseline:} Similar to FORGE but does not use force observations, dynamics randomization, or an excessive force penalty. However, it is trained with success prediction so that meaningful episode durations can be reported.

\textbf{Ablations:} In addition to the baselines, we also ablate each of the main components of {\sysName} for our sim-to-real analysis: Force (\textit{No Force}), Dynamics Randomization (\textit{No DR}), and Excessive Force Penalty (\textit{No FP}).
For the \textit{FORGE (No FP)} model, which ablates the contact penalty reward term, we evaluate using two P-gain levels.
Note that \textit{FORGE (No FP)} results are not reported for nut threading as we found that the nut always slipped out of the gripper.

\section{Results and Discussion}

\subsection{Baseline Comparisons}

\begin{table*}
    \vspace{1em}
    \centering
    \scriptsize
    \rowcolors{2}{gray!25}{white}
    \begin{tabular}[b]{ c|cc|cc|ccc }
      \hline
      & \multicolumn{2}{|c}{Episode} 
      & \multicolumn{2}{|c}{Force} & \multicolumn{3}{|c}{Early Termination} \\
      \hline
     \rowcolor{gray!75} 8mm Peg & Success Rate $\uparrow$ & Duration (s) $\downarrow$ & $F_{mean}$ (N) $\downarrow$ & $F_{max}$ (N) $\downarrow$ & Precision $\uparrow$ & Recall $\uparrow$ & Delay (s) $\downarrow$\\
     \hline 
     
FORGE  & 0.84 (0.05) & 2.82 (0.20) & 5.51 (0.24) & 12.84 (0.37) & 1.00 (0.0) & 1.00 (0.0) & 2.19 (0.08) \\
\hline
FORGE (No Force)  & 0.82 (0.06) & 3.18 (0.36) & 7.09 (0.35) & 14.16 (0.39) & 0.59 (0.08) & 0.81 (0.07) &  4.12 (0.37) \\
FORGE (No DR)  & 0.91 (0.04) & 2.03 (0.19) & 6.28 (0.24) & 13.08 (0.36) & 1.00 (0.0) & 0.98 (0.02) &  2.51 (0.04) \\
FORGE (No FP, 400kp)  & 0.64 (0.07) & 3.21 (0.35) & 6.94 (0.13) & 11.94 (0.24) & 0.83 (0.07) & 0.92 (0.05) &  2.85 (0.40) \\
FORGE (No FP, 600kp)  & 0.71 (0.07) & 2.88 (0.35) & 10.66 (0.15) & 16.58 (0.32) & 0.91 (0.05) & 0.97 (0.03) &  2.40 (0.22) \\
\hline
Baseline  & 0.64 (0.07) & 2.35 (0.27) & 11.81 (0.21) & 17.93 (0.41) & 0.97 (0.03) & 1.00 (0.0) &  2.74 (0.06) \\
IndustReal  & 0.82 (0.06) & 3.41 (0.24) & 9.45 (0.14) & 21.15 (0.26) & N/A & N/A &  6.59 (0.24) \\
     \hline 
     \hline

\rowcolor{gray!75} Medium Gear & Success Rate $\uparrow$ & Duration (s) $\downarrow$ & $F_{mean}$ (N) $\downarrow$ & $F_{max}$ (N) $\downarrow$ & Precision $\uparrow$ & Recall $\uparrow$ & Delay (s) $\downarrow$\\
     \hline 
     
FORGE  & 0.98 (0.02) & 3.14 (0.39) & 7.95 (0.11) & 15.10 (0.45) & 0.95 (0.03) & 1.00 (0.0) & 3.20 (0.28) \\
\hline
FORGE (No Force)  & 0.93 (0.04) & 3.06 (0.29) & 8.49 (0.23) & 14.68 (0.39) & 0.60 (0.08) & 1.00 (0.0) &  5.96 (0.70) \\
FORGE (No DR)  & 0.87 (0.05) & 3.42 (0.50) &  7.15 (0.20) & 13.94 (0.36) & 0.90 (0.05) & 1.00 (0.0) & 3.04 (0.24) \\
FORGE (No FP, 400kp)  & 0.82 (0.06) & 3.57 (0.26) & 6.52 (0.14) & 10.97 (0.24) & 1.00 (0.0) & 1.00 (0.0) & 2.87 (0.16) \\
FORGE (No FP, 600kp)  & 0.73 (0.07) & 3.08 (0.29) & 9.48 (0.23) & 15.73 (0.30) & 0.94 (0.04) & 0.97 (0.03) &  3.91 (0.39) \\
\hline
Baseline  & 0.69 (0.07) & 2.90 (0.37) & 11.67 (0.45) & 18.29 (0.40) & 0.90 (0.05) & 0.97 (0.03) &  4.68 (0.41) \\
IndustReal  & 0.87 (0.05) & 8.44 (0.61) & 9.80 (0.16) & 20.48 (0.26) & N/A & N/A &  6.56 (0.61) \\
     \hline
\hline

    \rowcolor{gray!75} M16 Nut & Success Rate $\uparrow$& Duration (s) $\downarrow$ & $F_{mean}$ (N) $\downarrow$ & $F_{max}$ (N) $\downarrow$ & Precision $\uparrow$ & Recall $\uparrow$ & Delay (s) $\downarrow$\\
     \hline 
FORGE & 0.69 (0.07) & 11.38 (0.47) & 7.82 (0.13) & 14.52 (0.22) & 0.74 (0.08) & 0.74 (0.08) &  6.54 (1.25) \\
\hline
FORGE (No Force)  & 0.40 (0.07) & 14.09 (1.11) & 8.34 (0.15) & 15.04 (0.17) & 0.33 (0.11) & 0.33 (0.11) &  11.48 (1.63) \\
FORGE (No DR)  &  0.56 (0.07) & 11.12 (0.14) & 7.65 (0.17) & 14.10 (0.24) & 0.72 (0.09) & 0.86 (0.08) &  8.10 (1.45) \\
\hline
Baseline  & 0.40 (0.07) & 11.19 (0.24) & 10.51 (0.51) & 17.34 (0.36) & 0.72 (0.11) & 0.93 (0.07) &  10.16 (1.69) \\
IndustReal  & 0.36 (0.07) & 22.27 (0.64) & 12.63 (0.31) & 22.34 (0.33) & N/A & N/A &  7.73 (0.64) \\
     \hline 

    \end{tabular}
\vspace{-1em}
\caption{\textbf{Baseline Comparison} {\sysName} is compared to baselines that are not force-aware (IndustReal \cite{tang2023industreal} and Baseline). It is also compared to ablations that do not observe force  (No Force), a force penalty (No FP), or dynamics rand (No DR). Evaluations are performed over a total of $855$ trials on the real robot ($45$ per row). Standard errors are included in parentheses.}
\label{tab:sim-success-rates}
\vspace{-2.5em}
\end{table*}

\textbf{(Q1) Does {\sysName} lead to more robust sim-to-real transfer?} \textbf{(Q2) Do {\sysName} policies have more desirable behavioural properties?}

Along with success rate, used to measure robustness for Q1, the following metrics are reported for Q2:
\begin{itemize}
\item Duration (s): For successful episodes, time to reach a successful state (independent of success prediction).
\item $F_{mean}, F_{max} (N)$: Forces experienced by the robot. 
\end{itemize}

Each reported metric represents $45$ trials spread across $5$ workspace locations for the fixed part, and $3$ position-estimation error levels ranging from $0-5mm$ (see Fig. \ref{fig:error-viz}). 
Similar randomization ranges were used as in simulation except for the in-hand part randomization where the part was centered in the gripper.
Results are reported in Table \ref{tab:sim-success-rates}.

One conclusion for Q1 is that {\sysName} outperformed the \emph{Baseline} method for all tasks and \emph{IndustReal} for both the gear meshing and nut threading tasks.
Ablations show that that the primary performance gains of {\sysName} come from including force observations and the excessive force penalty.
Although dynamics randomization did not  significantly affect success rate, we later show it is important for robustness across controller gains.

Examining the behavioural metrics for Q2, we notice that {\sysName} used less force than both baselines and had significant improvements in trial durations when compared to \emph{IndustReal}.
During experiments, we observed {\sysName} led to gentler interactions between the parts (see accompanying video).
The reduced force produced by this policy was especially helpful for the M16 Nut which was more susceptible to slipping than the peg or gear.%

For {\sysName}, the main failure cases occurred when there was high position estimation error (above $1\sigma$ of the training noise, see next section).
Parts got stuck on each other (peg insertion) or the nut was rotated before alignment with the bolt, causing the threads to miss.
However, we found training unstable with noise above $\sigma=2.5mm$.
Adopting a curriculum or adding additional sensing modalities (e.g., tactile sensors or wrist-cameras) may help address this.

\subsection{Noise Analysis}

\begin{figure}
    \centering
    \vspace{1em}
    \includegraphics[width=0.8\linewidth]{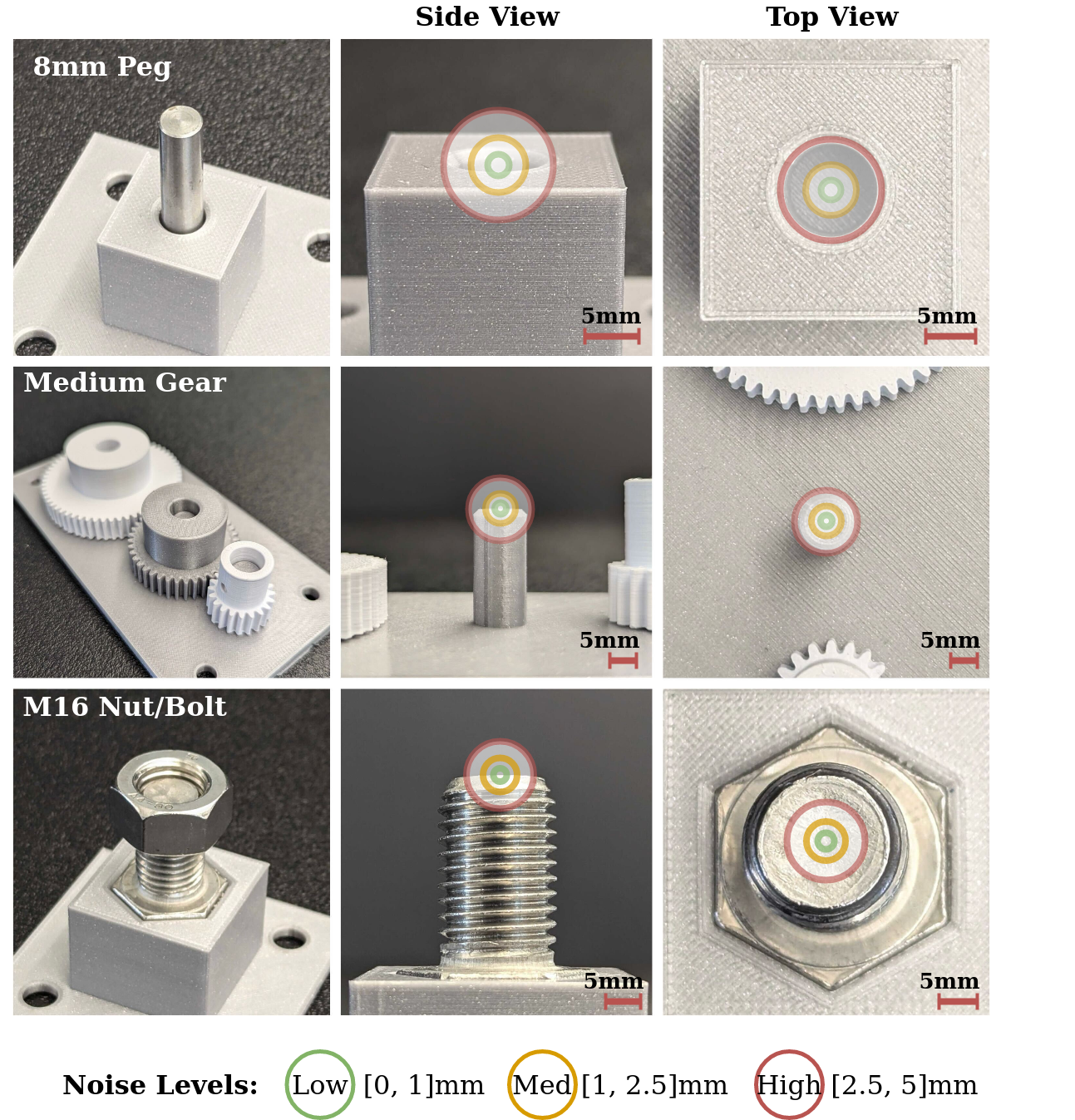}
    \vspace{-1em}
    \caption{\textbf{Perception Error} For each task, we visualize what the different position estimation errors look like overlaid on the fixed part.}
    \label{fig:error-viz}
    \vspace{-0em}
\end{figure}

\begin{figure}
    \centering
    \vspace{0.5em}
    \includegraphics[width=1.0\linewidth]{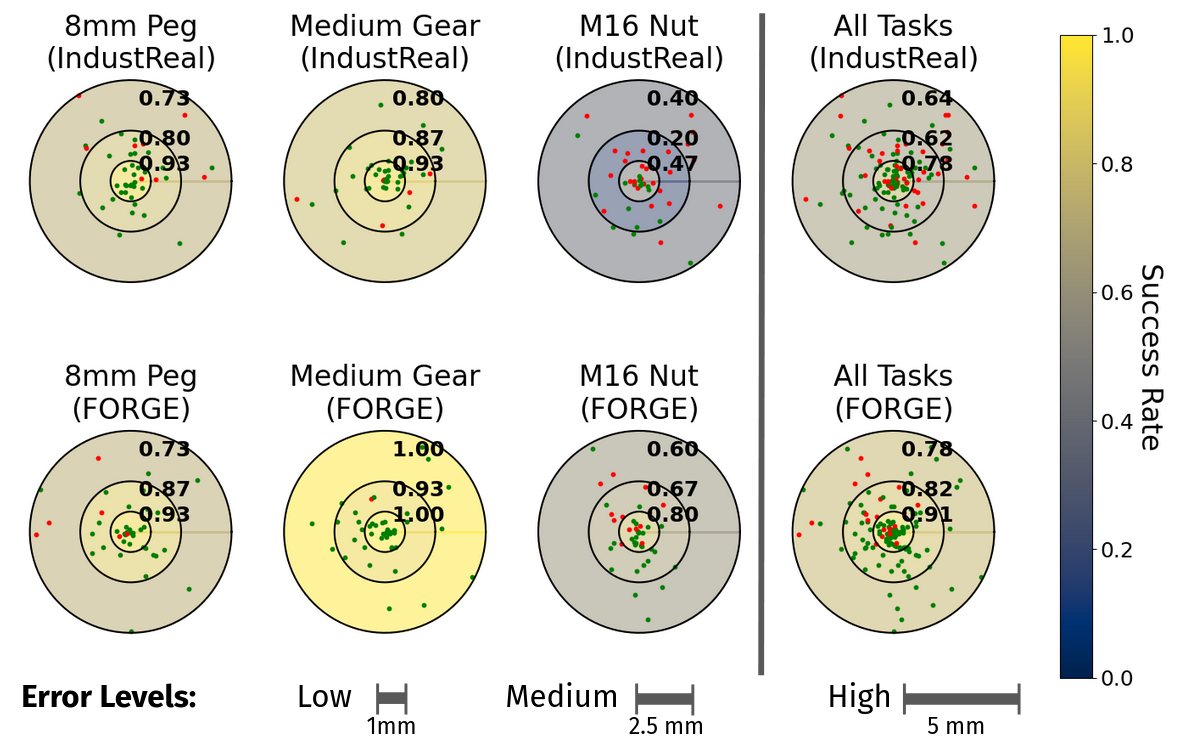}
    \vspace{-2.em}
    \caption{\textbf{Noise Analysis} Performance broken down by level of position error. Each subplot is a planar representation of the error levels where each ring corresponds to low (0-1mm), medium (1-2.5mm), and high (2.5-5mm) error. Success rate, stated in black text, is also represented by the shade of the corresponding ring. Dots represent x-y noise samples for successful (green) and failed (red) trials. {\sysName} results in good performance across tasks even with high error levels.}
    \label{fig:noise-analysis}
    \vspace{-1.0em}
\end{figure}

We next aim to answer \textbf{(Q3): How is policy performance, in terms of success rate, affected by position-estimation error?}
We use the same trials from the previous section, but show a breakdown of the results across different error levels.
During each trial, artificial perception error was added to the fixed part's position (calibrated as described in Section \ref{sec:exp-robot}\footnote{Adding artificial noise allows us to better characterize performance across error level compared to a perception system whose bias and variance can be difficult to estimate and control.}).
A third of the trials fell in each of the three considered error levels (see Fig. \ref{fig:error-viz}): Low (0-1mm), Medium (1-2.5mm), and High (2.5-5mm).
We considered $3D$ position error by sampling a perturbation vector with a radius uniformly sampled in the desired error range and a direction uniformly sampled from the unit-sphere.

Figure \ref{fig:noise-analysis} visualizes the performance of \emph{IndustReal} vs. {\sysName} policies at different noise levels.
Each subplot is a 2D representation of how much x-y error there was for each trial (z-dimension error not visualized).
Each point corresponds to either a successful (green) or unsuccessful (red) trial (only $xy$ corrindates of the error vector are visualized as dots).
The color of the ring represents the success rate at the corresponding error levels (increasing outwards).

Although performance is comparable for the peg insertion task, {\sysName} outperformed \emph{IndustReal} for the gear meshing and nut threading tasks at all noise levels.
This demonstrates that force is a useful modality to robustly recover from larger amounts of position estimation error.
Performance generally degraded with error $>2.5mm$ which is beyond $1\sigma$ of the observation noise added in simulation.
With high error, the effects of contact are more pronounced because the robot may need to search longer before the task is complete.

\subsection{Force Analysis}

\begin{figure}
    \centering
    \includegraphics[trim={0 0cm 0 0cm},clip,width=1.0\linewidth]{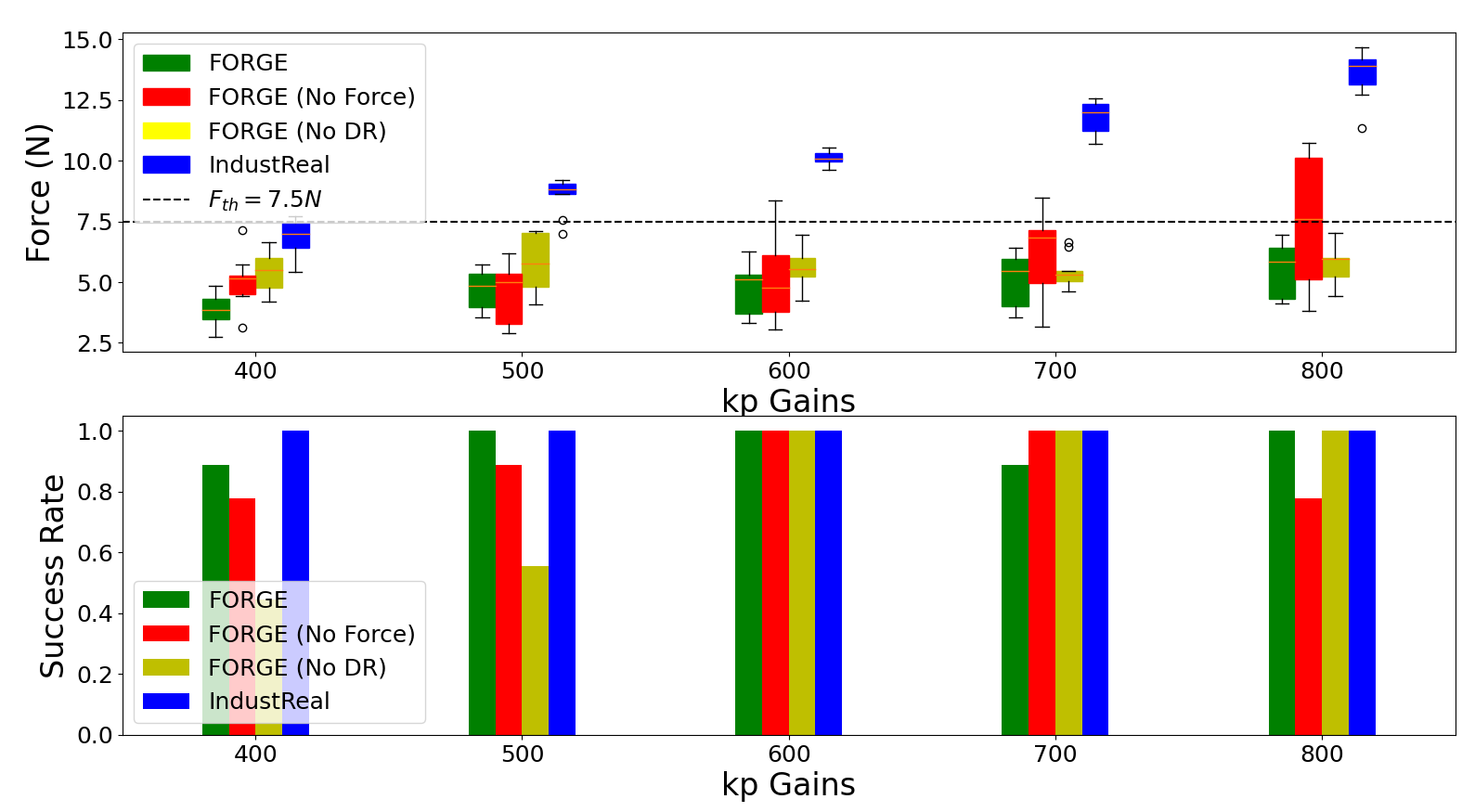}
    \vspace{-2em}
    \caption{\textbf{Gains Analysis} (180 trials, 8mm Peg) With force sensing, {\sysName} can achieve robust success rates (bottom) across varying controller gains at deployment time. Even with different gains, force sensing allows the policy to modulate its actions to achieve low contact forces (top). }
    \label{fig:gains-analysis}
    \vspace{-1.5em}
\end{figure}

Next, we investigate how {\sysName} limits forceful interactions. \textbf{(Q4) How important is the excessive-force penalty for safe interactions?} \textbf{(Q5) Can {\sysName} limit the applied force without extensive controller tuning?}

\textbf{Excessive-Force Penalty (Q4):} 
In Table \ref{tab:sim-success-rates}, we compare to an ablation, \emph{No FP}, that was trained without the excessive-force penalty of {\sysName} (but still used force observations and dynamics randomization).
We used the same evaluation procedure as for {\sysName} but deployed with two different controller gains (we chose values at the lower and middle of the gain randomization range).
We found that policies deployed with the lower gains achieved similar average forces to {\sysName} while those deployed with higher gains naturally experienced more force.
Both policies had lower success rates than {\sysName} which was deployed with controller gains at the middle of the randomization range.

\textbf{Gains Robustness (Q5):}
To measure how robust {\sysName} is to controller gains, we performed an additional experiment where we varied the gains at deployment time and measured success rate.
We compare {\sysName} to \emph{IndustReal} and multiple ablations.
The experiment was carried out for the $8mm$ peg task at a single workspace location, with medium position estimation error and limited initial-state randomization.
We considered $5$ proportional gain levels across the randomization range (corresponding to an $8N$ range in the maximum force the controller could apply) and each condition was evaluated $9$ times ($3$ runs per checkpoint).

In Fig. \ref{fig:gains-analysis}, we see that {\sysName} achieves high success rates while respecting the force threshold across a wide range of controller gains.
However, performance is less consistent without force observations or dynamics randomization.
In Fig. \ref{fig:gains-analysis} (top), we use a box plot to show the spread of $F_{mean}$ across the $9$ trials of each condition.
The dotted line shows the deployment force-threshold: $F_{th}=7.5N$.
We see that when the force observation was included, contact force was consistently low across gains.
However, without force observations, the spread of forces across episodes was high, often exceeding the threshold at higher gains.
Similarly, the force exerted by \emph{IndustReal} policies increased with controller gains.
As \emph{IndustReal} is not force-aware, achieving desired forceful properties requires tuning controller gains.
Overall, these results highlight the importance of force sensing to enable the policy to effectively modulate the contact force.

\subsection{Success Prediction Analysis}
\label{sec:results-ft}
To evaluate success prediction, we ask: \textbf{(Q6) Does success prediction,  trained in simulation, transfer to the real world?} \textbf{(Q7) Can success prediction be used to tune the force-threshold for tasks that require forceful insertion?}

\textbf{Sim-to-Real Transfer (Q6):} 
To measure the effective of success prediction, we report additional metrics for each of the trials in Table \ref{tab:sim-success-rates}:
\begin{itemize}
\item Early Term. Precision: The fraction of early-terminated trials that were actually successful.
\item Early Term. Recall: The fraction of successful trials which were terminated correctly with $a^{ET} > p_{term}$.
\item Early Term. Delay (s): For successful episodes, how long after success occurred did the policy terminate.
\end{itemize}
Results show that success prediction transferred well to the real world.
The termination method correctly identified successes (high precision and recall) and worked best when using force observations for all tasks.
We also see that delay times are shortest when using force observations.
This shows the benefit of force for sensing task completion: when the gear has been fully meshed or the nut threads successfully engaged.
Using success prediction also leads to shorter delays than \emph{IndustReal} which uses a fixed duration.

\begin{figure}
    \centering
    \vspace{1em}
    \includegraphics[width=0.475\linewidth,trim={1cm 0cm 1.25cm 1.5cm},clip]{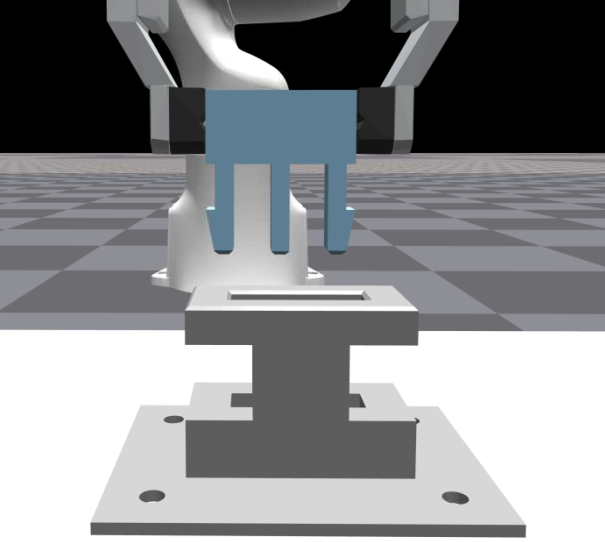}
    \includegraphics[width=0.475\linewidth,trim={0cm 0cm 0cm 2.25cm},clip]{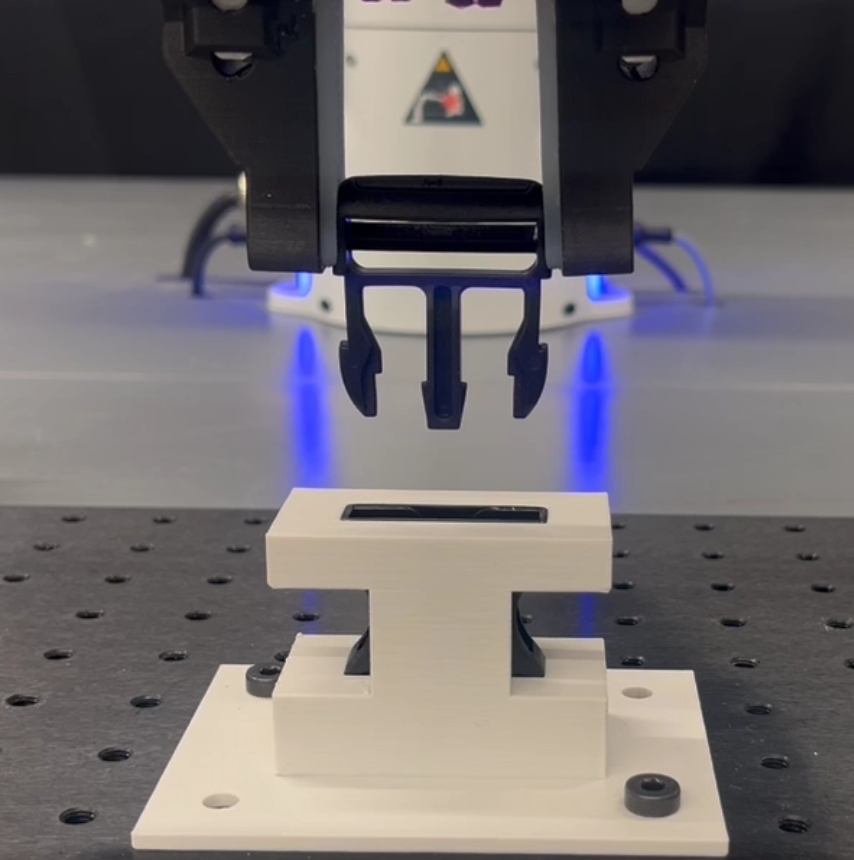}
    \vspace{-0.5em}
    \caption{\label{fig:snapfit} Simulated (left) and real (right) parts for a snap-fit insertion task which requires $15N$ of force to succeed. See the video for policy execution.}
    \vspace{-2em}
\end{figure}

\textbf{Force Threshold Tuning (Q7):} To evaluate the utility of \emph{success prediction} for force threshold tuning, we introduced a new \emph{snap-fit} task (see Fig. \ref{fig:snapfit}).
In simulation, the snap-fit buckle was implemented with torsion springs for each of the clips. 
The stiffness of the springs was randomized to vary the amount of force needed for insertion.
We also ensured the robot's gains and force-threshold were randomized such that success was possible.
In the real world, we used a snap-fit buckle that required $15N$ of force for insertion.
We report real world results from running the automatic tuning procedure described in Sec. \ref{forge-term}.
The initial force threshold was set to $7.5N$ and increased by $5N$ each policy execution until the policy predicted success.
No initial state randomization or noise were added for these experiments.

Out of $10$ trials for the complete tuning procedure (each consisting of up to $3$ policy executions with increasing force thresholds), the tuning procedure succeeds $8$ times while successful insertion occurred $10$ times.
The two failures were instances where the insertion succeeded, but the policy failed to predict success.
Success occurred on the third execution $9/10$ times (it once occurred on the second trial), meaning the policy generally respected the force threshold (success should not occur until the force threshold exceeds $15N$).
Furthermore, of the $29$ policy executions, the success prediction by the policy was correct $27$ times. 

We also evaluated the task using a sufficiently high threshold and on a larger initial state distribution, similar to what was used during training in simulation. Here the success rate dropped to $6/10$, which we attribute to an unstable grasp leading to part slippage during contact.
This reveals a limitation of having a single force threshold: for certain tasks, the optimal force threshold may vary depending on the phase of the task. For example, low forces are required until the buckle is aligned with the socket, only then is it safe to use high forces.

\subsection{Multi-Stage Assembly}
\label{sec:results-gearbox}

\begin{figure}
    \centering
    \vspace{1em}
    \includegraphics[width=0.95\linewidth]{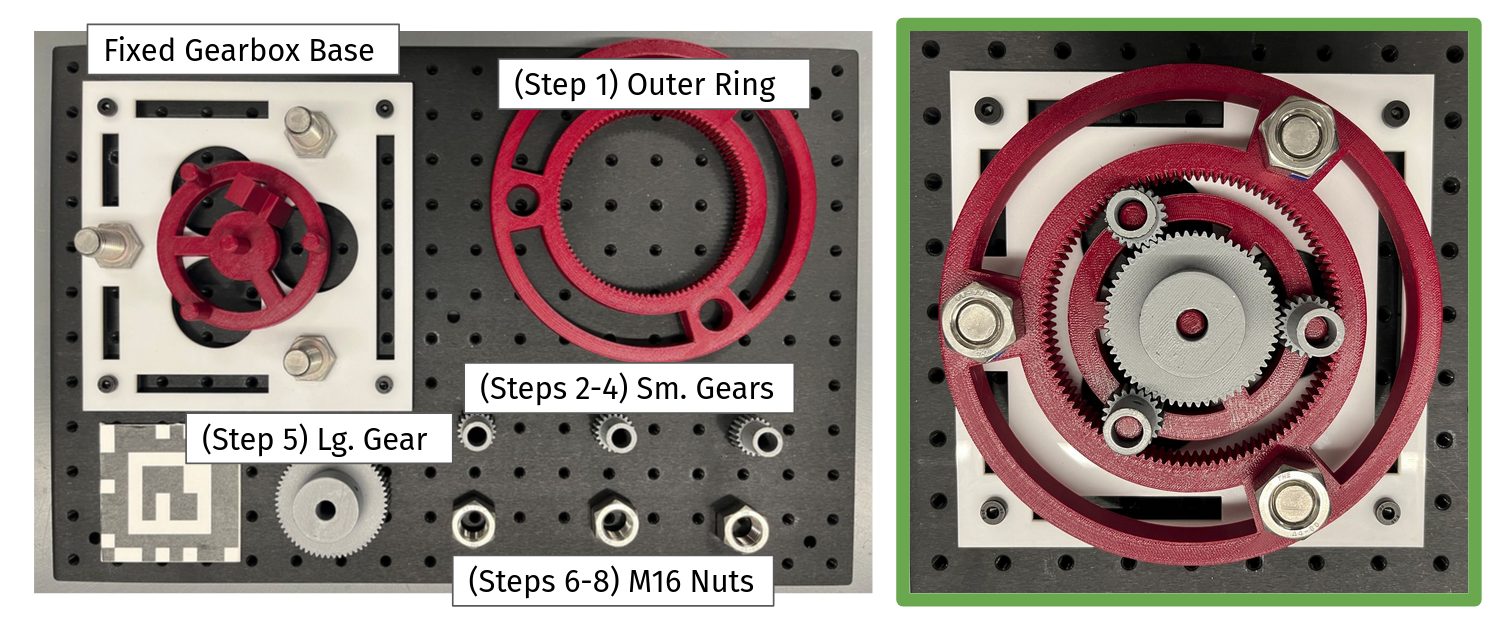}
    \vspace{-1.25em}
    \caption{\label{fig:gearbox2}{\sysName} policies enable a robot to complete long-horizon tasks such as assembling a planetary gearbox (from initial state [left] to goal state [right, enlarged]).}
    \vspace{-1.5em}
\end{figure}

To culminate this work, we show that {\sysName} enables the multi-stage assembly of a planetary gearbox using a simple perception system (see Fig. \ref{fig:gearbox2} for the initial and final states).
We assume the assembly sequence is known \textit{a priori} and train {\sysName} policies for \emph{Small Gear}, \emph{Large Gear}, and \emph{M16 Nut} tasks.
We additionally introduce a new \emph{Ring Insertion} task, which must also be robust to orientation estimation noise such that the three bolts align with the holes in the outer ring.
Successfully assembling the planetary gearbox requires executing $8$ contact-rich primitives.

We ran $5$ trials resulting in the following success rates: Ring Insertion ($5/5$), Small Gear ($15/15$), Large Gear ($3/5$), M16 Nut ($15/15$).
Early terminations saved on average $65s$ in a single trial compared to executing policies for a fixed duration.
Overall, the complete assembly succeeded in $3/5$ trials where the failures correspond to the large gear insertion (which has to align the teeth of three already inserted small gears).
Please see the accompanying video for a demonstration of the multi-stage assembly and App. \ref{app:gearbox} for more experimental details.

\section{Related Work}
Assembly tasks typically involve mating parts with tight clearances and detailed geometries  \cite{xu2019compare, jia2018survey}. 
Various approaches have been proposed to handle pose uncertainty in such tasks.
Mechanically, \textit{remote centers of compliance} \cite{drake1978using} or chamfers can mitigate small misalignments. 
Compliant control \cite{lozano1984automatic} and strategies such as spiral search \cite{chhatpar2001search, newman2001interpretation} have also be used for insertion.
These strategies typically consider low noise levels and are task-specific.

\textbf{Real World Reinforcement Learning}
A large body of work focuses on learning assembly tasks directly on the real-robot.
Learning on the robot side-steps the \emph{sim-to-real} gap by using data (and contact-interactions) from the same distribution expected at deployment.
These works typically address problem of data efficiency by leveraging demonstrations \cite{abu2018force, davchev2022ral, Luo-RSS-21, vecerik2019icra, luo2024serl} or using model-based approaches \cite{luo2019reinforcement, fan2019icra, lee2020icra, luo2018iros}.
To ensure excessive forces are not exceeded during training, these papers typically use control methods designed to be safe \cite{luo2024serl, inoue2017iros, lee2020tro}.

\textbf{Sim-to-Real Transfer:}
Learning directly in simulation is often preferable for robot safety, increased task variability, and access to privileged state.
With advancements in RL and parallelizable simulation \cite{todorov2012mujoco, coumans2021, drake, makoviychuk2021isaac}, there has been much interest in \emph{sim-to-real} transfer for complex control problems.
Of note include legged locomotion \cite{hwangbo2019learning, agarwal_legged_2022, margolis_rapid_2022, rudin_learning_nodate} and in-hand manipulation \cite{allshire2021transferring, akkaya2019solving, handa2022dextreme}.

Recent advances in contact-rich simulation has enabled efficient simulation of assembly tasks  \cite{lan2022affine, macklin2020local, narang2022factory, yoon2022fast}.
However, as discussed throughout the paper, the key challenge becomes the sim-to-real gap: how can we \emph{safely} and \emph{successfully} deploy policies that were trained in simulation?

Although \emph{system identification} is a principled approach to minimize the \emph{sim-to-real} gap \cite{ljung1998system}, it is often time-consuming and difficult to apply to contact-rich tasks \cite{acosta2022validating, guo2023ldcc}.
Instead, \emph{dynamics randomization} randomizes parameters such as part friction/stiffness \cite{beltran2020variable, peng_sim--real_2018, oren2021iros, apolinarska2021con, hebecker2021aim}, controller gains \cite{jin2021acc, peng_sim--real_2018}, or F/T observation scale \cite{jin2021acc, zhang2024bridging}.
Even with randomization, excessive forces can occur when deployed.
An expert can tune the controller gains at deployment or choose an action-space that is safe by design \cite{tang2023industreal, zhang2020modular, vuong2021icra}.
Gains can also be adapted online via optimization \cite{zhang2023efficient} or an explicit \emph{gain-tuning} model \cite{zhang2024bridging}.

Similar to FORGE, other works have proposed to use a force-threshold \cite{beltran2020variable, martin2019variable, hebecker2021aim}.
These works have a fixed threshold during training which is often very large to primarily prevent damage (e.g., $40N$).
However, especially with small parts, slip can occur with much lower contact forces.
Most similar to FORGE, \cite{beltran2020variable} introduces a method to specify the \emph{desired} interaction force at deployment time.

Most prior work focus on insertion-style tasks. 
We show how the combined application of a force-threshold and dynamics randomization can lead to robust sim-to-real transfer for a range of tasks, including the complicated nut-threading task.
Prior work on sim-to-real for nut-threading \cite{son_sim--real_2020} focused on large parts ($M48$ nuts) that were fixed to the gripper.
In addition, we show these techniques are applicable for sim-to-real transfer of success prediction.

\textbf{Success Prediction:}
Previous \emph{sim-to-real} approaches execute policies for a fixed duration \cite{tang2023industreal}.
Instead, we would like to terminate once success is achieved.
For some tasks, success can be manually specified from sensor data \cite{pinto2016supersizing,wen2022you}.
For others, a classifier can be learned from visual data \cite{su2018learning, fu2018variational}.
However, for contact-rich tasks, visual and proprioceptive data alone may be insufficient to determine success \cite{5584452}. %
In such cases, the robot can execute actions to verify success \cite{kroemer2021review}. 
Previous work learns a separate policy to check success \emph{after} task execution \cite{huang2022training}. 
Instead, we jointly trained a policy to predict success \textit{during} task execution.

\section{Conclusion}
In conclusion, we present {\sysName}, a force-aware method to train robust sim-to-real policies with pose estimation uncertainty.
{\sysName} uses a force threshold and dynamics randomization to learn \emph{safe} exploration behaviours, enabling successful policy execution with up to $5mm$ of position estimation error.
In addition, {\sysName} can predict task success, allowing efficient policy execution and force threshold tuning.
In future work, we plan to investigate torque sensing for more efficient search strategies.
We also believe research in \emph{real-to-sim} will help automatically tune simulation models for more adaptive behaviours.

\section*{ACKNOWLEDGMENT}

The authors thank the Seattle Robotics Lab and the Robust Robotics Group for their valuable feedback.

\bibliographystyle{IEEEtran}
\bibliography{IEEEabrv,example}

\clearpage
\appendix

\subsection{Randomization}
\label{app:randomization}
All randomization ranges are reported in Table \ref{tab:exp-params}.
In addition to the dynamics randomization described in the text, we also randomize the initial state distribution and observation noise.

\textbf{Initial State Randomization}: At the start of an episode, we randomize the position of the fixed part, the relative pose of the hand above the fixed part, and the relative position of the held part in the gripper (where the default position has the top of the held part aligned with the bottom of the gripper).

\textbf{Observation Randomization}:
In simulation, the position of the fixed asset is randomized once per episode by adding Gaussian noise.
Independent Gaussian noise is added to each observation at every timestep (except velocity, where positional noise is propagated through finite differencing).

\subsection{Reward}
\label{app:reward}

\subsubsection{Keypoint Reward}
Here we describe the keypoint reward in more details.
Keypoint distance is calculated as: $d^{kp}_t(p^{held}_t, p^{fixed}) = ||k^{held}_t - k^{targ}||$.
The target keypoints, $k^{targ}$, represent the desired position of the held part, while $k^{held}_t$ represent its current position.
We use a logistic kernel as in \cite{allshire2021transferring} to transform keypoint distances into a bounded reward: $\mathcal{K}_{a,b}(d_{kp}) = (e^{-ax} + b + e^{ax})^{-1}$.
The kernel can be tuned to be sensitive to distances at different scales using parameters $a$ and $b$ (see Table \ref{tab:exp-params}).

Using a single kernel parameterization was not sufficient for the nut-threading task due to small geometry.
Different phases of the task require motion at different scales.
For example, initial placement of the nut on the bolt requires movement ranging from $0-2cm$.
However, lowering the nut by the final thread changes the position by $<0.1cm$.
Instead, we propose a \textit{coarse-to-fine} keypoint reward. 
The final reward is a sum of: 
(1) A \emph{coarse reward} directing the arm towards the tip of the fixed part
and; (2) a \emph{fine reward} incentivizing more detailed motion once the arm is close to the part.
These are implemented using different parameters for the logistic kernel, 
\begin{equation}
    R_{kp}(p^{fixed}, p^{held}_t) = \mathcal{K}^{coarse}_{a^c, b^c}(d^{kp}_t) + \mathcal{K}^{fine}_{a^f, b^f}(d^{kp}_t).
\end{equation}
Parameters for each task can be found in Table \ref{tab:exp-params}.

\subsubsection{Task Success}

Each task defines success based on the relative positions between the held and fixed parts (Table \ref{tab:exp-params} shows \emph{Success Dist.} as the distance between the top of the fixed part and bottom of the held part when success is achieved):
\begin{itemize}
    \item \textit{Peg Insertion}: The bottom of the peg is within $1mm$ of the base of the socket (equivalently, $24mm$ below the top of the socket).
    \item \textit{Gear Meshing}: The bottom of the gear is within $1mm$ of the base of the gear plate (equivalently, $19mm$ below the tip of the gear peg).
    \item \textit{Nut Threading}: The $M16$ nut is lowered a quarter thread (corresponding to $2.5mm$ below the tip of the bolt, as the first thread is chamfered). 
\end{itemize}
For all tasks, success also requires the parts to be laterally centered.

\begin{table}
    \footnotesize
    \rowcolors{2}{gray!25}{white}
    \begin{tabular}[b]{ p{2cm}@{\hskip 0.25cm}p{1.75cm}@{\hskip 0.25cm}p{1.75cm}@{\hskip 0.25cm}p{1.75cm} }
        \hline
        \rowcolor{white} \multicolumn{4}{c}{Initial State Randomization} \\ 
        \rowcolor{white} Parameter & \multicolumn{3}{c}{All Tasks} \\
        \hline
        Fixed: $x,y,z$ & \multicolumn{3}{l}{$[0.55,0.65]m, [-0.05,0.05]m, [0.0,0.1]m$} \\
        Hand: $x,y$ (rel) & \multicolumn{3}{l}{$[-2,2]cm, [-2,2]cm$} \\
        Held: $x,y$ (rel) & \multicolumn{3}{l}{$[-3,3]mm, [0,0]mm$} \\ 
        \hline
        \rowcolor{white} Parameter & $8mm$ Peg & Medium Gear & $M16$ Nut \\ 
        \hline
        Hand: $z$ (rel) & $[3.7, 5.7]cm$ & $[2.5, 4.5]cm$ & $[0.5, 2.5]cm$ \\
        Hand: $yaw$  & $[-45, 45]\degree$ &$[-45, 45]\degree$ & $[-120, -90]\degree$\\
        Held: $z$ (rel) & $[14, 20]mm$ & $[12, 15]mm$&  $[10, 16]mm$\\
            
        \hline\hline
        \rowcolor{white} \multicolumn{4}{c}{Observation Randomization} \\ 
        \rowcolor{white} Parameter & $8mm$ Peg & Medium Gear & $M16$ Nut \\
        \hline
        Pos-Est Noise & $2.5mm$ & $2.5mm$ & $2.5mm$ \\
        Force Noise & $1N$ & $1N$ & $1N$ \\
        EE-Pos. Noise & $0.25mm$ & $0.25mm$ & $0.25mm$ \\
        
        \hline\hline
         \rowcolor{white} \multicolumn{4}{c}{Dynamics Randomization} \\ 
        \rowcolor{white} Parameter & $8mm$ Peg & Medium Gear & $M16$ Nut \\
        \hline
        Part Friction & $[0.5, 1.0]$ & $[0.38, 0.75]$ & $[0.1, 0.38]$\\
        Controller Gains & $[400, 800]$ & $[400, 800]$ & $[400, 800]$\\
        Action Scale: $\lambda$ & $[1.6, 2.5]cm$ & $[1.6, 2.5]cm$ & $[1.6, 2.5]cm$ \\
        Dead Zone & $[0,5]N$ & $[0,5]N$ & $[0,5]N$ \\
        Force Threshold & $[5, 10]N$ & $[5, 10]N$ & $[5, 10]N$\\
        
        \hline\hline
        \rowcolor{white} \multicolumn{4}{c}{Reward Specification} \\ 
        \rowcolor{white} Parameter & $8mm$ Peg & Medium Gear & $M16$ Nut \\
        \hline
        Coarse $(a^c, b^c)$ & $(50, 2)$ & $(50, 2)$ & $(100, 2)$ \\
        Fine: $(a^f, b^f)$ & $(100, 0)$ & $(100, 0)$ & $(500, 0)$ \\
        Contact-Pen: $\beta$ & $0.2$ & $0.05$ & $0.05$ \\
        Success Dist. & $24mm$ & $19mm$ & $2.5mm$ \\
        Place Dist. & $2.5mm$ & $2mm$& $2.5mm$\\
        Episode Length & $150$ ($10s$) & $300$ ($20s$) & $450$ ($30s$) \\
        \hline
    
    \end{tabular}
\caption{\label{tab:exp-params} Simulation parameters used to train {\sysName} policies.} 
\vspace{-2em}
\end{table}

\subsection{IndustReal Baseline}
\label{app:industreal}
IndustReal \cite{tang2023industreal} proposes a series of techniques for sim-to-real transfer of contact-rich tasks:
\begin{itemize}
\item Simulation Aware Policy Updates (SAPU, sim): Penalize a policy when its actions lead to interpenetration.
\item SDF Rewards (sim): Compute rewards based on signed distances between target and current asset point clouds.
\item Sampling Based Curriculum (SBC, sim): Training progresses in difficulty by descreasing the fraction of environments that start near the goal state.
\item Policy Level Action Integrator (PLAI, real): A method to smooth actions and reduce steady-state error.
\end{itemize}
The authors extensively evaluate the system on two contact-rich tasks: peg insertion and gear meshing.

\textbf{Key Differences:} Although IndustReal presents strong results, it struggles for tasks that require delicate manipulation such as nut-threading.
IndustReal policies do not use force observations and are not trained to avoid excessive forces.
In simulation, IndustReal policies use small gains to avoid large forces by default (resulting in maximum applied forces of $3N$).
On the real robot, as our results show, a policy's forceful behaviour is largely determined by the controller gains used at deployment.
In the IndustReal work, these gains were set to large values, resulting in maximum achievable forces of $15N$ or higher.
In addition to IndustReal policies causing part slip, we noticed another common failure case for the nut-threading task. The arm would rotate before the nut was in contact with the bolt, causing failed thread alignment.
We posit that force is a useful modality to detect when parts are aligned. 
This is otherwise difficult to do when the pose estimation error is too large.

\textbf{Implementation:} Beyond the key components of each framework, there are additional differences between FORGE and the original IndustReal implementation. 
To make the comparison as informative and fair as possible, we used our own version of IndustReal with the following changes:
\begin{itemize}
\item Policy Frequency: Like FORGE, our version of IndustReal used a policy inference rate of $15Hz$ (compared to the original $60Hz$). This also required increasing the PLAI action scales by a factor of $4$.
\item Policy Network and Training Hyperparameters: We use the same network structure and training parameters for all methods.
\item Pose Estimation Noise: Our implementation uses FORGE's noise model in simulation ($\sigma=2.5mm$). Although this is higher than the $unif(-1mm, 1mm)$ sampling distribution originally used in IndustReal, we found the policies still trained reliably.
\end{itemize}

\textbf{Nut Threading:} In our work, we also consider the nut-threading task which was not considered in the original IndustReal work.
We directly applied our IndustReal implementation but found that the SDF-reward function was poorly suited to learn successful threading.
This is because there are only small SDF distances between partially threaded and unthreaded nuts with the same orientation.
Instead, for this task only, we used the coarse-to-fine keypoint reward discussed in App. \ref{app:reward}.
All other components of IndustReal were kept the same.

\subsection{Early-Termination}
\label{app:early-term}
\begin{figure}
    \centering
    \vspace{1em}
     \includegraphics[trim={0 0 0 0},clip,width=1.0\linewidth]{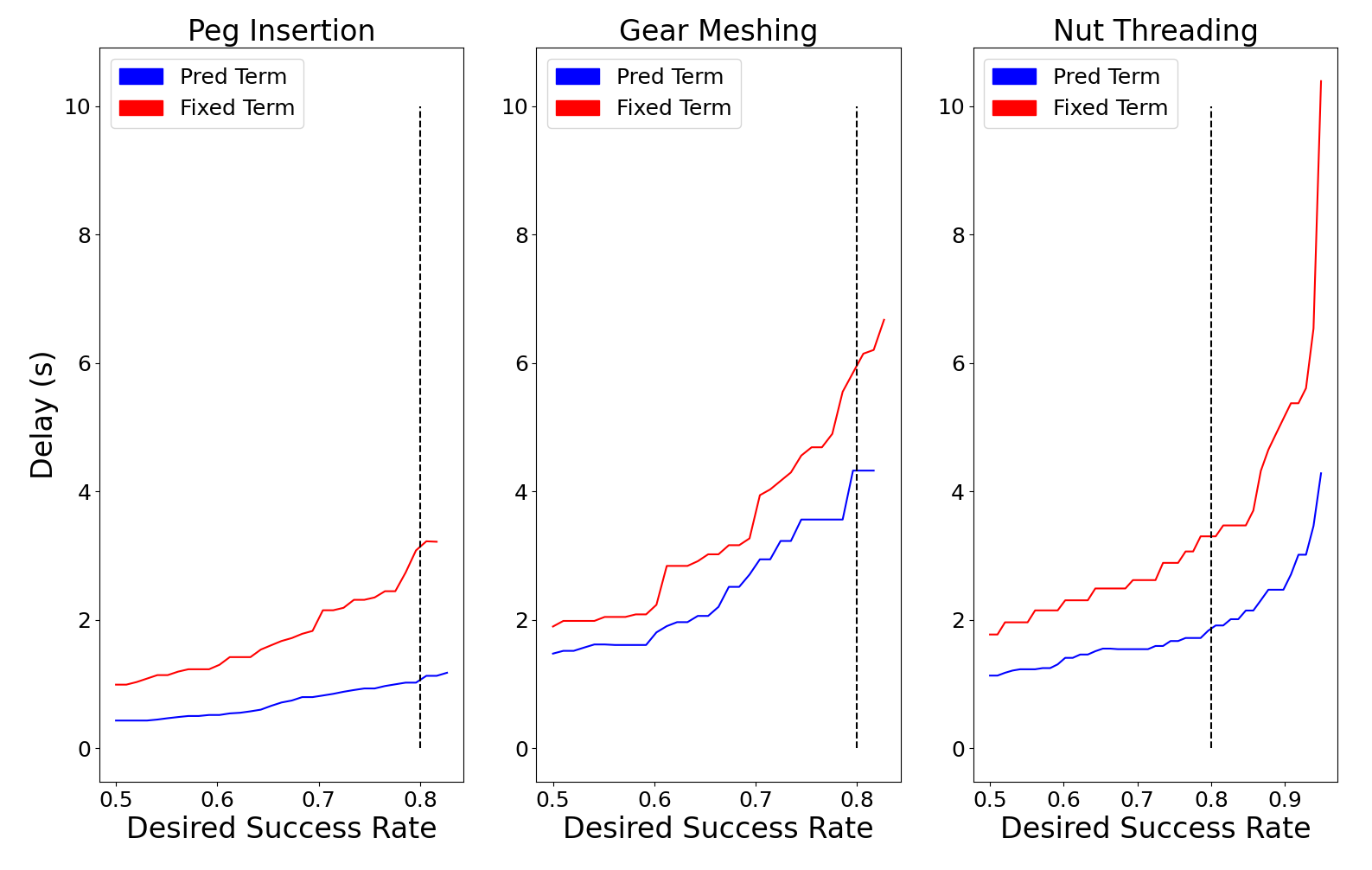}
    \vspace{-2.5em}
    \caption{\label{fig:success-prediction-analysis} \textbf{Success Prediction Analysis} Relationship between Delay Time and Success Rate for two early termination methods (generated by varying each method's respective parameter: $T$ or $p_{term}$). The \emph{Pred Term} method leads to lower delays than the \emph{Fixed Term} method, especially at higher success rates. The vertical line shows a $0.8$ success rate.}
    \vspace{-1.5em}
\end{figure}

Making decisions based on the success prediction action, $a_t^{ET}$, involves choosing a threshold, $p_{term}$.
In this section, we investigate the trade-offs made when choosing this threshold.

We use \emph{Delay Time (s)} to capture efficiency (lower values are better). 
Delay time measures the time between when success occurred and when the episode was terminated ($a_t^{ET} > p_{term}$).
We compare the proposed method (Pred Term) to a standard termination method that stops the policy after a fixed duration, $T$ (Fixed Term).
Each method has a parameter that can be tuned to produce a different success rate (fraction of episodes that are successful when terminated). 
However, this will introduce a trade-off with delay time: 
\begin{itemize}
\item Fixed Term ($T$): Waiting too long is inefficient while terminating too early will harm success rates.
\item Pred Term ($p_{term}$): A high threshold can cause extra delay while a low threshold can affect success rate.
\end{itemize}

Fig. \ref{fig:success-prediction-analysis} is a simulated analysis that shows the relationship between \emph{Delay Time} and \emph{Success Rate} for each method.
Each line was generated by measuring the success rate and corresponding delay time across a fine discretization of each method's termination parameter.
These were then sorted by success rate and plotted.\footnote{Similar to an ROC plot, but higher areas above the curve are better.}
As a practitioner, one could choose a desired success rate and find the resulting delay.

Across all tasks, we see that the \emph{Fixed Term} method leads to longer delays, especially at higher success rates (we plot a vertical line to show the $0.8$ success rate).
The early termination action, $a^{ET}$, allows for dynamic episode lengths, leading to high success rates with smaller delay times.

\subsection{Snap-fit Task}
\label{app:snapfit}
\textbf{Simulation Model:} For forceful insertion, we consider a snap-fit task.
In the real world, these parts typically have clips that deform for successful insertion.
Because \emph{Factory} \cite{narang2022factory} uses a rigid-body simulator, we approximate deformation by using a spring model on the snap-fit clips.
Varying the stiffness of this spring changes the amount of force necessary for insertion.

\textbf{Training Details:} At deployment time, we assume the amount of force required for insertion is unknown.
To ensure a single policy can solve snap-fit tasks with varying force requirements, we randomize both the gains of the torsion spring and the force threshold when training in simulation.
We ensure the force threshold is higher than the necessary force required for insertion.
For this environment only, we found it was helpful for the policy to have noisy proportional gains of the controller as input.

\subsection{Planetary Gearbox}
\label{app:gearbox}
For the planetary gearbox, we trained policies for the following tasks: Ring Insertion, Small Gear Meshing, Large Gear Meshing, and M16 Nut Threading.

\textbf{Gear Tasks}: The gearbox requires insertion of three small gears, each with one abutting gear, and one large gear with three abutting gears.
In simulation, the small and large gear meshing tasks had one abutting gear each.
This is similar to deployment for the small gear which achieved a high success rate ($15/15$).
However, when the large gear is deployed, it needs to mesh with the three already inserted small gears.
This is much harder than how the policy was trained and could be a cause of the performance drop for this task ($3/5$).
Note these statistics come from five executions of the entire planetary gearbox assembly (and hence a different number of trials per gear size).

\begin{figure}
    \centering
    \includegraphics[width=0.75\linewidth]{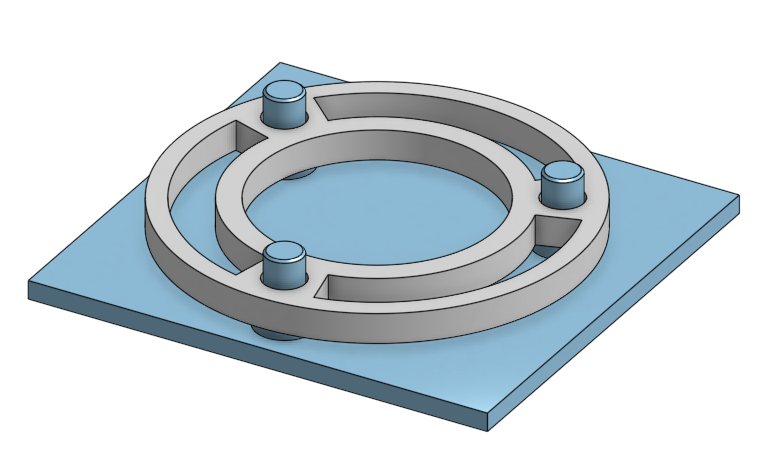}
    \vspace{-1.75em}
    \caption{\label{fig:gearbox} Simulated assets for the ring insertion task. The ring gear (grey) is inserted onto the gearbox plate (blue).}
    \vspace{-1.5em}
\end{figure}

\textbf{Ring Insertion}: The outer ring gear must be inserted onto the three bolts of the gearbox base.
We designed simulation assets for the corresponding parts (see Fig. \ref{fig:gearbox}) and trained a policy using the {\sysName} framework.
We assume there is small yaw error on the ring ($< 5 \degree$) during training.
Success is defined as having the ring gear placed close to the base ($<2mm$ displacement) and all three bolt holes aligned.

\textbf{Gearbox Design}: Note, we also designed a ``lock'' for the gear carrier which is removed by the robot after the small gears are inserted.
This ensures a fixed base during the small gear insertions (see video).

\textbf{Policy Selection}: All policies were trained using the {\sysName} framework including force observations.
We trained one policy per task without any additional checkpoint selection procedure.
The M16 policy was chosen as the best policy from our main evaluation.
For the gearbox experiments only, we selected high control stiffness for the roll and pitch dimensions of the impedance controller, as the policy does not generate actions for these degrees of freedom.

\textbf{Task Execution}: To pick up the held parts, we assume a known grasp location which was predetermined (with small noise from placement error). 
However, the location of the corresponding fixed parts were estimated from the \emph{IndustReal} perception system \cite{tang2023industreal}.
Grasping and movement to the initial state for policy execution was performed with a standard position controller.
No additional artificial noise or initial-state randomization was added for the gearbox experiments.

\textbf{Perception:} The perception system from IndustReal \cite{tang2023industreal} assumes the $z$-position of parts are known and uses a Mask-RCNN model \cite{he2017mask} to estimate bounding boxes from which planar locations can be backed out.
We retrained the Mask-RCNN model using data we collected.
Perception errors are largely caused by extrinsic calibration and bounding box prediction errors.

\subsection{Generalization across Part Geometry}
\label{app:generalization}
\begin{figure*}
    \centering
    \includegraphics[trim={0 2cm 0 1cm},clip,width=0.3\linewidth]{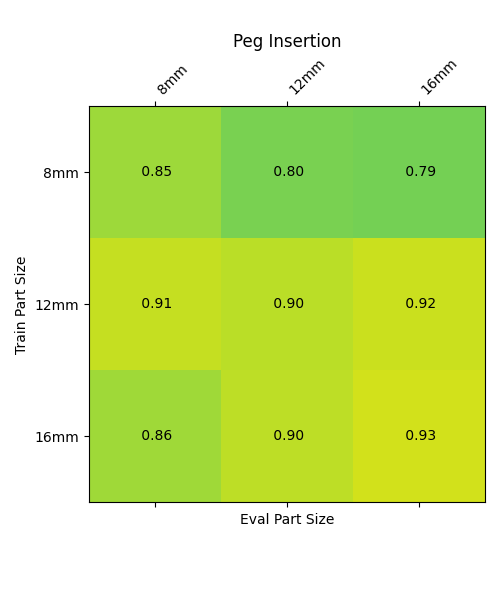}
    \includegraphics[trim={0 2cm 0 1cm},clip,width=0.3\linewidth]{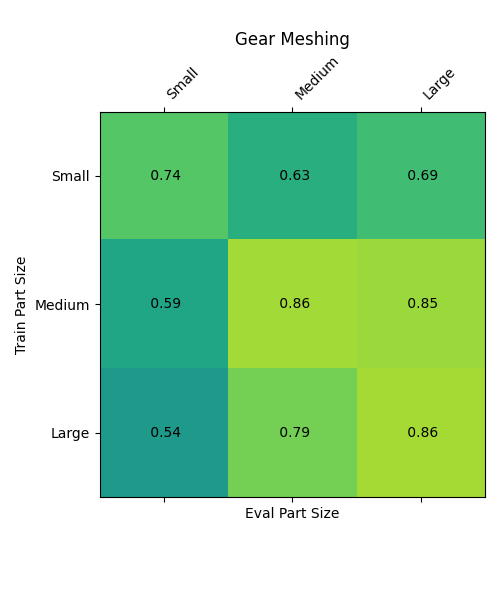}
    \includegraphics[trim={0 2cm 0 1cm},clip,width=0.3\linewidth]{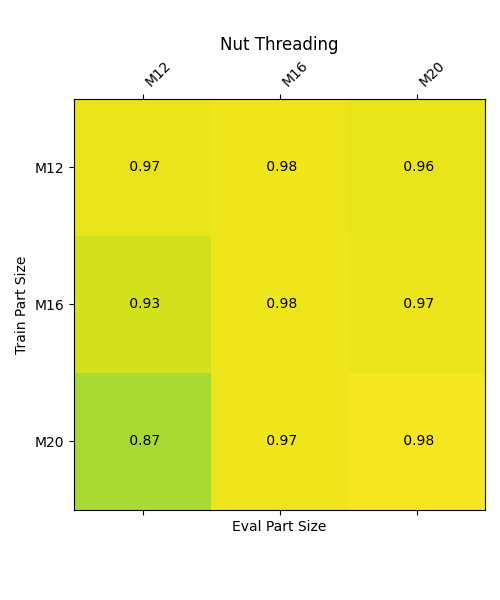}
    \vspace{-1em}
    \caption{\textbf{Part Size Generalization} Specialist policies trained on a single part size tend to generalize to other part sizes. Each cell aggregates success rates from $3$ policies trained with different random seeds and evaluated $128$ times each.}
    \label{fig:app-generalization}
\end{figure*}

For our real robot experiments, we focused on a single part size for each task.
In this section, we show that FORGE policies achieve similar simulated performance across part sizes.
To do so, we train and evaluate specialist policies for three part sizes for each task.
\begin{itemize}
    \item Peg Insertion:  $8mm$, $12mm$, $16mm$
    \item Gear Meshing: Small, Medium, Large
    \item Nut Threading: $M12$, $M16$, $M20$
\end{itemize}

To assess policy generalization, we also evaluate policy performance in simulation for the part sizes they were not trained on.
The results in Fig. \ref{fig:app-generalization} show the success rates of this evaluation.
Policies achieve high performance on the part size for which they were trained.
In most cases, policies also generalize to other part sizes.
The case with the worst generalization is ``Train on Medium/Large Gear'' and ``Evaluate on Small Gear''. This can be explained because of the significant geometry differences between the parts. The small gear has a much smaller base, so a search strategy that would work for the larger gears, would cause the small one to fall off the peg. 
In future work, we are hopeful we can train a single policy per task which generalizes across multiple part geometries by randomizing geometry in simulation.

\end{document}